\documentclass[12pt]{article}

\usepackage{pgffor}
\usepackage{listings}
\usepackage{authblk}

\usepackage{hyperref,amsmath,latexsym,xspace}
\usepackage{tikz}
\usetikzlibrary{tikzmark, arrows}
\usetikzlibrary{positioning,arrows.meta}
\usetikzlibrary{matrix}
\usetikzlibrary{fit}
\usetikzlibrary{decorations.pathreplacing}

\usepackage{amssymb}
\usepackage{booktabs}
\usepackage{float}
\usepackage{fancyvrb}
\usepackage{listings}
\makeatletter
\let\old@lstKV@SwitchCases\lstKV@SwitchCases
\def\lstKV@SwitchCases#1#2#3{}
\makeatother
\usepackage{lstlinebgrd}
\makeatletter
\let\lstKV@SwitchCases\old@lstKV@SwitchCases

\lst@Key{numbers}{none}{%
    \def\lst@PlaceNumber{\lst@linebgrd}%
    \lstKV@SwitchCases{#1}%
    {none:\\%
     left:\def\lst@PlaceNumber{\llap{\normalfont
                \lst@numberstyle{\thelstnumber}\kern\lst@numbersep}\lst@linebgrd}\\%
     right:\def\lst@PlaceNumber{\rlap{\normalfont
                \kern\linewidth \kern\lst@numbersep
                \lst@numberstyle{\thelstnumber}}\lst@linebgrd}%
    }{\PackageError{Listings}{Numbers #1 unknown}\@ehc}}
\makeatother

\usepackage{xcolor}
\usetikzlibrary{arrows}

\definecolor{codegreen}{rgb}{0,0.6,0}
\definecolor{codegray}{rgb}{0.5,0.5,0.5}
\definecolor{codepurple}{rgb}{0.58,0,0.82}
\definecolor{backcolour}{rgb}{0.95,0.95,0.92}

\usepackage[multiple]{footmisc}

\usepackage[round,authoryear]{natbib}
\def\bx{{\boldsymbol x}}

\def\bh{{\boldsymbol h}}

\begin{document}
  \tikzset{
    arr/.style={draw opacity=1,
      decoration={markings,mark=at position 1
        with {\arrow[scale=2]{>}}},
      postaction={decorate}, shorten >=0.4pt,
    }
  }
  \title{A Step-by-step Introduction to the Implementation of Automatic Differentiation}

\author[1]{Yu-Hsueh Fang$^*$}
\author[1]{He-Zhe Lin$^*$}
\author[1]{Jie-Jyun Liu}
\author[1,2]{Chih-Jen Lin}
\affil[1]{National Taiwan University}
\affil[ ]{{\tt\{d11725001, r11922027, d11922012\}@ntu.edu.tw}}
\affil[ ]{{\tt cjlin@csie.ntu.edu.tw}}
\affil[2]{Mohamed bin Zayed University of Artificial Intelligence}
\affil[ ]{{\tt chihjen.lin@mbzuai.ac.ae}}
\def\thefootnote{*}\footnotetext{These authors contributed equally to this work}
\def\thefootnote{\arabic{footnote}}
% https://www.csie.ntu.edu.tw/~cjlin/papers/libsvm.pdf
% https://www.csie.ntu.edu.tw/~cjlin/papers/liblinear.pdf
% \editor{}
\date{\today}
\maketitle

\setlength{\baselineskip}{18pt}
\begin{abstract}
  Automatic differentiation is a key component in deep learning.
  This topic is well studied and excellent surveys such as~\cite{AGB18a} have been available to clearly describe the basic concepts.
  Further, sophisticated implementations of automatic differentiation are now an important part of popular deep learning frameworks.
  However, it is difficult, if not impossible, to directly teach students the implementation of existing systems due to the complexity.
  On the other hand, if the teaching stops at the basic concept, students fail to sense the realization of an implementation.
  For example, we often mention the computational graph in teaching automatic differentiation, but students wonder how to implement and use it.
  In this document, we partially fill the gap by giving a step by step introduction of implementing a simple automatic differentiation system.
  We streamline the mathematical concepts and the implementation.
  Further, we give the motivation behind each implementation detail, so the whole setting becomes very natural.
\end{abstract}

\section{Introduction}\label{sec1:Intro}
In modern machine learning, derivatives are the cornerstone of numerous applications and studies.
The calculation often relies on automatic differentiation, a classic method for efficiently and accurately calculating derivatives of numeric functions.
For example, deep learning cannot succeed without automatic differentiation.
Therefore, teaching students how automatic differentiation works is highly essential.
\par Automatic differentiation is a well-developed area with rich literature.
Excellent surveys including \cite{SC94a}, \cite{MBB00a}, \cite{AGB18a} and \cite{CCM19a} review the algorithms for automatic differentiation and its wide applications.
In particular, \cite{AGB18a} is a comprehensive work focusing on automatic differentiation in machine learning.
Therefore, there is no lack of materials introducing the concept of automatic differentiation.
\par On the other hand, as deep learning systems now solve large-scale problems, 
it is inevitable that the implementation of automatic differentiation becomes highly sophisticated.
For example, in popular deep learning systems such as PyTorch \citep{AP17a} and Tensorflow \citep{MA16a}, at least thousands of lines of code are needed.
Because of this, many places of teaching automatic differentiation for deep learning stop at the basic concepts.
Then students fail to sense the realization of an implementation.
For example, we often mention the computational graph in teaching automatic differentiation, but students wonder how to implement and use it.
In this document, we aim to partially fill the gap by giving a tutorial on the basic implementation.\
\par In recent years, many works\footnote{\url{https://towardsdatascience.com/build-your-own-automatic-differentiation-program-6ecd585eec2a}}$^{,}$\footnote{\url{https://sidsite.com/posts/autodiff}}$^{,}$\footnote{\url{https://mdrk.io/introduction-to-automatic-differentiation-part2/}}$^{,}$\footnote{\url{https://github.com/dlsyscourse/lecture5/blob/main/5_automatic_differentiation_implementation.ipynb}}$^{,}$\footnote{\url{https://github.com/karpathy/micrograd}}
have attempted to discuss the basic implementation of automatic differentiation.
However, they still leave room for improvement.
For example, some are not self-contained -- they quickly talk about implementations without connecting to basic concepts.
Ours, which is very suitable for the beginners, has the following features:
\begin{itemize}
  \item We streamline the mathematical concepts and the implementation.
  Further, we give the motivation behind each implementation detail, so the whole setting becomes very natural.
  \item We use the example from~\cite{AGB18a} for the consistency with past works.
  Ours is thus an extension of \cite{AGB18a} into the implementation details. 
  \item We build a complete tutorial including this document, slides and the source code at\\ \url{https://www.csie.ntu.edu.tw/~cjlin/papers/autodiff/}.
\end{itemize}

\section{Automatic Differentiation}
% Description
There are two major modes of automatic differentiation.
In this section, we introduce the basic concepts of both modes.
Most materials in this section are from~\cite{AGB18a}.
We consider the same example function
\[
  y = f(x_1, x_2) = \log x_1 + x_1x_2 - \sin x_2.
\]
\subsection{Forward Mode}
First, we discuss the forward mode.
Before calculating the derivative, let us check how to calculate the function value.
Assume that we want to calculate the function value at $(x_1,x_2)=(2,5)$.
Then, in the following table, we have a forward procedure.
\begin{center}
  \begin{tabular}{l l l l}
    \tikzmark{a}& $x_1$ & & $=2$ \\
    & $x_2$ &  &  $=5$  \\
    \cmidrule{2-4}
    & $v_{1}$ & $= \log{x_1}$ &  $=\log{2}$  \\
    & $v_{2}$ & $= x_1\times x_2$ &  $=2\times5$  \\
    & $v_{3}$ & $= \sin{x_2}$ &  $=\sin{5}$  \\
    & $v_{4}$ & $= v_{1}+v_{2}$ &  $=0.693 + 10$  \\
    & $v_{5}$ & $= v_{4}-v_{3}$ &  $=10.693 + 0.959$  \\
    \cmidrule{2-4}
    \tikzmark{b}& $y$ & $=v_{5}$ &  $=11.652$ 
  \end{tabular}
  \begin{tikzpicture}[overlay,remember picture, shorten >=-3pt]
    \draw[->,>=triangle 60,thick] ([shift={(0ex,1.5ex)}]pic cs:a) -- (pic cs:b) ;
    \end{tikzpicture}
\end{center}
We use variables $v_i$ to record the intermediate outcomes.
First, we know $\log$ function is applied to $x_1$.
Therefore, we have $\log(x_1)$ as a new variable called $v_1$.
Similarly, there is a variable $v_2$, which is $x_1\times x_2$. 
Each $v_i$ is related to a simple operation.
The initial value of this member is zero when a node is created.
In the end, our function value at $(2,5)$ is $y=v_5$.
As shown in the table, the function evaluation is decomposed into a sequence of simple operations.
We have a corresponding computational graph as follows:
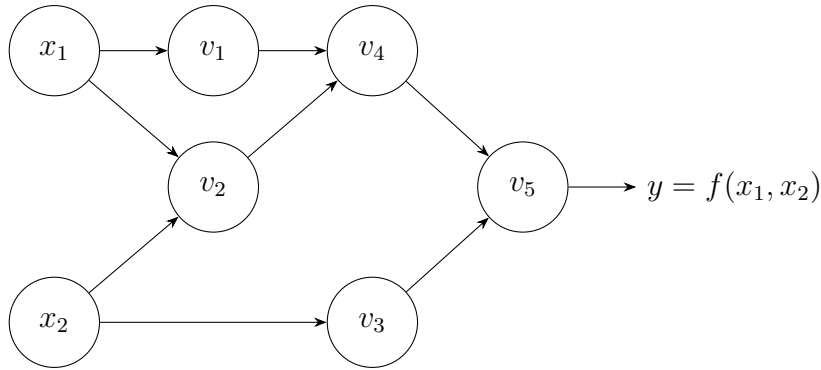
\begin{center}
\begin{tikzpicture}[auto,vertex/.style={draw,circle},minimum size=1.2cm]
\node[vertex] (b) at (0,0) {$x_1$};
\node[vertex,right=0.9cm of b] (c) {$v_{1}$};
\node[vertex,right=0.9cm of c] (d) {$v_{4}$};
\node[vertex,below=2.4cm of b] (g) {$x_2$};
\node[vertex,below=0.6cm of c] (h) {$v_2$};
\node[vertex,below=2.4cm of d] (i) {$v_{3}$};
\node[vertex,right=2.9cm of h] (j) {$v_5$};
\node[right=0.9cm of j] (k) {$y=f(x_1, x_2)$};
%\draw[->,>=Stealth] ([xshift=-1cm]b.west) node[xshift=-8]{$x_1$} -- (b);  
%\draw[->,>=Stealth] ([xshift=-1cm]g.west) node[xshift=-8]{$x_2$} -- (g);
\path[-{Stealth[]}]
(b) edge (c)
(c) edge (d)
(b) edge (h)
(g) edge (h)
(h) edge (d)
(d) edge (j)
(g) edge (i)
(i) edge (j)
(j) edge (k);
\end{tikzpicture}
\end{center}
Because calculating both $v_{1}$ and $v_{2}$ needs $x_1$, $x_1$ has two links to them in the graph.
The following graph shows all the intermediate results in the computation.
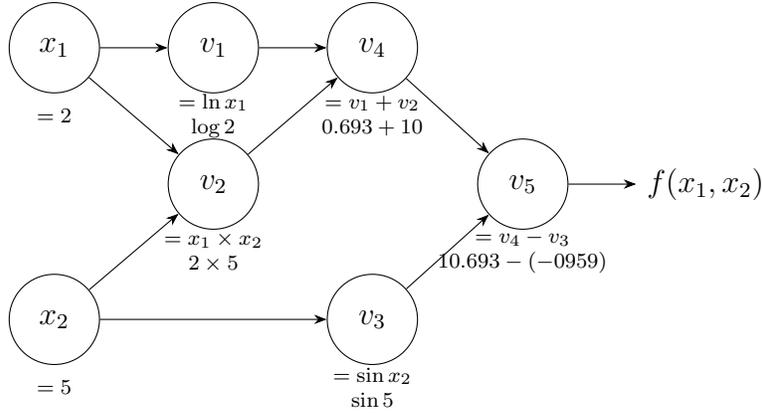
\begin{center}
  \begin{tikzpicture}[
    auto,vertex/.style={draw,circle},minimum size=1.2cm,
    every label/.append style={font=\scriptsize,below,yshift=-0.9cm}
    ]
    \node[vertex,label={$=2$}] (b) at (0,0) {$x_{1}$};
    \node[vertex,right=0.9cm of b,
      label={[align=center]$= \ln x_{1}$\\$\log 2$}] (c) {$v_{1}$};
    \node[vertex,right=0.9cm of c,
      label={[align=center]$= v_1 + v_2$\\ $0.693+10$}] (d) {$v_{4}$};
    \node[vertex,below=2.4cm of b,
      label={[align=center]$= 5$}] (g) {$x_{2}$};
    \node[vertex,below=0.6cm of c,
      label={[align=center]$= x_{1}\times x_{2}$\\$2\times 5$}] (h) {$v_2$};
    \node[vertex,below=2.4cm of d,
      label={[align=center]$= \sin{x_{2}}$\\$\sin{5}$}] (i) {$v_{3}$};
    \node[vertex,right=2.9cm of h,
      label={[align=center]$= v_4 - v_3$\\$10.693-(-0959)$}] (j) {$v_5$};
    \node[right=0.9cm of j] (k) {$f(x_1, x_2)$};
    \path[-{Stealth[]}]
    (b) edge (c)
    (c) edge (d)
    (b) edge (h)
    (g) edge (h)
    (h) edge (d)
    (d) edge (j)
    (g) edge (i)
    (i) edge (j)
    (j) edge (k);
  \end{tikzpicture}
\end{center}

% \begin{center}
%   \begin{tikzpicture}[
%   auto,vertex/.style={draw,circle},minimum size=1.2cm,
%   every label/.append style={font=\scriptsize,below,yshift=-0.8cm}
%   ]
%     \node[vertex,label=2] (b) at (0,0) {$x_1$};
%     \node[vertex,right=0.9cm of b,label=$\log{2}$] (c) {$v_{1}$};
%     \node[vertex,right=0.9cm of c,label=0.693+10] (d) {$v_{4}$};
%     \node[vertex,below=2.4cm of b,label=5] (g) {$x_2$};
%     \node[vertex,below=0.6cm of c,label=$2\times 5$] (h) {$v_2$};
%     \node[vertex,below=2.4cm of d,label=$\sin{5}$] (i) {$v_{3}$};
%     \node[vertex,right=2.9cm of h,label=10.693-(-0959)] (j) {$v_5$};
%     \node[right=0.9cm of j] (k) {$y=f(x_1, x_2)$};
%     % \draw[->,>=Stealth] ([xshift=-1cm]b.west) node[xshift=-8]{$x_1$} -- (b);
%     % \draw[->,>=Stealth] ([xshift=-1cm]g.west) node[xshift=-8]{$x_2$} -- (g);
%     \path[-{Stealth[]}]
%       (b) edge (c)
%       (c) edge (d)
%       (b) edge (h)
%       (g) edge (h)
%       (h) edge (d)
%       (d) edge (j)
%       (g) edge (i)
%       (i) edge (j)
%       (j) edge (k);
%   \end{tikzpicture}
%   \end{center}
The computational graph tells us the dependencies of variables.
Thus, from the inputs $x_1$ and $x_2$ we can go through all nodes for getting the function value $y=v_5$ in the end.

Now, we have learned about the function evaluation.
But remember, we would like to calculate the derivative.
Assume that we target at the partial derivative ${\partial y}/{\partial x_1}$.
Here, we denote
\begin{equation*}
\dot{v} = \frac{\partial v}{\partial x_1}
\end{equation*}
as the derivative of the variable $v$ with respect to $x_1$.
The idea is that by using the chain rule, 
we can obtain the following forward derivative calculation to eventually get $\dot{v}_5 = \partial f / \partial x_1$.
\begin{center}
\begin{tabular}{l l l l}
  \tikzmark{c}& $\dot{x}_{1}$ & $= \partial x_1 / \partial x_1$ & $=1$ \\
  & $\dot{x}_{2}$ & $= \partial x_2 / \partial x_1$ &  $=0$  \\
  \cmidrule{2-4}
  & $\dot{v}_{1}$ & $= \dot{x}_{1}/ x_1$ &  $=1/2$  \\
  & $\dot{v}_{2}$ & $= \dot{x}_{1}\times x_2 + \dot{x}_{2}\times x_1$ &  $=1\times 5+ 0\times 2$  \\
  & $\dot{v}_{3}$ & $= \dot{x}_{2}\times\cos{x_2}$ &  $=0\times\cos{5}$  \\
  & $\dot{v}_{4}$ & $= \dot{v}_{1}+\dot{v}_{2}$ &  $=0.5 + 5$  \\
  & $\dot{v}_{5}$ & $= \dot{v}_{4}-\dot{v}_{3}$ &  $=5.5 - 0$  \\
  \cmidrule{2-4}
  \tikzmark{d}& $\dot{y}$ & $=\dot{v}_{5}$ &  $=5.5$
\end{tabular}
\begin{tikzpicture}[overlay,remember picture, shorten >=-3pt]
\draw[->,>=triangle 60,thick] ([shift={(0ex,1.5ex)}]pic cs:c) -- (pic cs:d) ;
\end{tikzpicture}
\end{center}
The table starts from $\dot{x}_{1}$ and $\dot{x}_{2}$, which are ${\partial x_1}/{\partial x_1}=1$ and ${\partial x_2}/{\partial x_1}=0$.
Based on $\dot{x}_{1}$ and $\dot{x}_{2}$, we can calculate other values.
For example, let us check the partial derivative $\partial{v_1}/\partial{x_1}$.
From
\begin{equation*}
  v_1 = \log x_1,
\end{equation*}
by the chain rule,
\begin{equation*}
  \dot{v}_{1} =  \frac{\partial v_1}{\partial x_1} \times \frac{\partial x_1}{\partial x_1} =  \frac{1}{x_1} \times \frac{\partial x_1}{\partial x_1} =  \frac{\dot{x}_{1}}{x_1}.
\end{equation*}
Therefore, we need $\dot{x}_1$ and $x_1$ for calculating $\dot{v}_{1}$ on the left-hand side.
We already have the value of $\dot{x}_{1}$ from the previous step (${\partial x_1}/{\partial x_1}=1$).
Also, the function evaluation gives the value of $x_1$.
Then, we can calculate $\dot{x}_{1}/x_1=1/2$.
Clearly, the chain rule plays an important role here.
The calculation of other $\dot{v}_i$ is similar.

\subsection{Reverse Mode}
Next, we discuss the reverse mode.
We denote
\begin{equation*}
  \bar{v} = \frac{\partial y}{\partial v}
\end{equation*}
as the derivative of the function $y$ with respect to the variable $v$.
Note that earlier, in the forward mode, we considered
\begin{equation*}
  \dot{v} = \frac{\partial v}{\partial x_1},
\end{equation*}
so the focus is on the derivatives of all variables with respect to one input variable.
In contrast, the reverse mode focuses on $\bar{v}={\partial y}/{\partial v}$ for all $v$, 
the partial derivatives of one output with respect to all variables.
Therefore, for our example, we can use $\bar{v}_i$'s and $\bar{x}_i$'s to get both ${\partial y}/{\partial x_1}$ and ${\partial y}/{\partial x_2}$ at once.
Now we illustrate the calculation of
\begin{equation*}
  \dfrac{\partial y}{\partial x_2}.
\end{equation*}
\par By checking the variable $x_2$ in the computational graph, we see that variable $x_2$ affects $y$ by affecting $v_2$ and $v_3$.
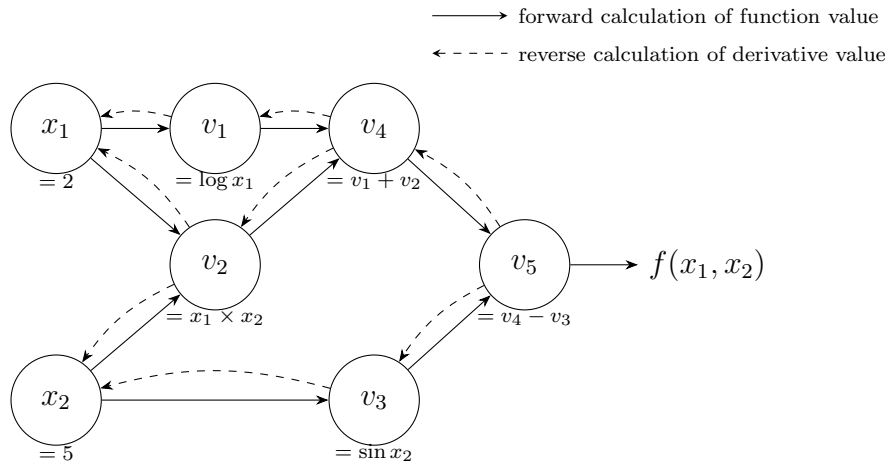
\begin{center}
  \begin{tikzpicture}[
    auto,vertex/.style={draw,circle},minimum size=1.2cm,
    every label/.append style={font=\scriptsize,below,yshift=-0.7cm}
    ]
    \node[vertex,label={$=2$}] (b) at (0,0) {$x_{1}$};
    \node[vertex,right=0.9cm of b,label={$= \log x_{1}$}] (c) {$v_{1}$};
    \node[vertex,right=0.9cm of c,label={$= v_1 + v_2$}] (d) {$v_{4}$};
    \node[vertex,below=2.4cm of b,label={$= 5$}] (g) {$x_{2}$};
    \node[vertex,below=0.6cm of c,label={$= x_{1}\times x_{2}$}] (h) {$v_2$};
    \node[vertex,below=2.4cm of d,label={$= \sin{x_{2}}$}] (i) {$v_{3}$};
    \node[vertex,right=2.9cm of h,label={$= v_4 - v_3$}] (j) {$v_5$};
    \node[right=0.9cm of j] (k) {$f(x_1, x_2)$};
    % \draw[->,>=Stealth] ([xshift=-1cm]b.west) node[xshift=-8, label=2]{$x_1$} -- (b);
    % \draw[->,>=Stealth] ([xshift=-1cm]g.west) node[xshift=-8, label=5]{$x_2$} -- (g);
    \path[-{Stealth[]}]
    (b) edge (c)
    (c) edge (d)
    (b) edge (h)
    (g) edge (h)
    (h) edge (d)
    (d) edge (j)
    (g) edge (i)
    (i) edge (j)
    (j) edge (k)
    (c) edge[bend right=15, dashed] node [left] {} (b)
    (d) edge[bend right=15, dashed] node [left] {} (c)
    (h) edge[bend right=15, dashed] node [left] {} (b)
    (h) edge[bend right=15, dashed] node [left] {} (g)
    (d) edge[bend right=15, dashed] node [left] {} (h)
    (j) edge[bend right=15, dashed] node [left] {} (d)
    (i) edge[bend right=15, dashed] node [left] {} (g)
    (j) edge[bend right=15, dashed] node [left] {} (i);
  % Legend
    \draw[->, >= stealth] (5,1.5) -- (6,1.5) node[right,font=\scriptsize] {forward calculation of function value};
    \draw[<-, dashed, >= stealth] (5,1) -- (6,1) node[right,font=\scriptsize] {reverse calculation of derivative value};
  \end{tikzpicture}
\end{center}This dependency, together with the fact that $x_1$ is fixed, means if we would like to calculate $\partial y/\partial x_2$, then it is equal to calculate
\begin{equation}\label{ypartialtov0}
  \frac{\partial  y}{\partial x_2}
  = \frac{\partial  y}{\partial v_2}
  \frac{\partial  v_2}{\partial x_2}
  +
  \frac{\partial  y}{\partial v_3}
  \frac{\partial  v_3}{\partial x_2}.
\end{equation}
We can rewrite Equation~\ref{ypartialtov0} as follows with our notation.
\begin{equation}
  \bar{x}_2
        = \bar{v}_2
      \frac{\partial  v_2}{\partial x_2}
      + \bar{v}_3
      \frac{\partial  v_3}{\partial x_2}.
\end{equation}
If $\bar{v}_2$ and $\bar{v}_3$ are available beforehand, all we need is to calculate ${\partial  v_2}/{\partial x_2}$ and ${\partial  v_3}/{\partial x_2}$.
From the operation between $x_2$ and $v_3$, we know that ${\partial v_3}/{\partial x_2}=\cos(x_2)$.
Similarly, we have ${\partial v_2}/{\partial x_2}=x_1$.
Then, the evaluation of $\bar{x}_2$ is done in two steps:
\begin{equation*}
  \begin{split}
    &
    \bar{x}_2 \leftarrow
            \bar{v}_3
    \frac{\partial  v_3}{\partial x_2}\\
&       \bar{x}_2 \leftarrow
      \bar{x}_2 +  \bar{v}_2
    \frac{\partial  v_2}{\partial x_2}.
  \end{split}
\end{equation*}
These steps are part of the sequence of a reverse traversal, shown in the following table.
\begin{small}
\begin{center}
  \begin{tabular}{l l l l l}
    \tikzmark{e}& $\bar{x}_1$  & & & $=5.5$ \\
    & $\bar{x}_2$ & & & $=1.716$  \\
    \cmidrule{2-5}
    &$\bar{x}_{1}$ &$=\bar{x}_{1}+\bar{v}_{1}\frac{\partial v_1}{\partial x_1}$ & $= \bar{x}_{1}+\bar{v}_1 / x_1$ &  $=5.5$  \\
    &$\bar{x}_{2}$& $=\bar{x}_{2}+\bar{v}_2\frac{\partial v_2}{\partial x_2}$ & $= \bar{x}_{2}+ \bar{v}_{2}\times x_1$ &  $=1.716$  \\
    &$\bar{x}_{1}$ &$=\bar{v}_{2}\frac{\partial v_2}{\partial x_1}$ & $=\bar{v}_{2}\times x_2$ &  $=5$\\
    &$\bar{x}_{2}$ &$=\bar{v}_{3}\frac{\partial v_3}{\partial x_2}$ & $= \bar{v}_{3}\times\cos{x_2}$ &  $=-0.284$\\
    &$\bar{v}_{2}$ &$=\bar{v}_{4}\frac{\partial v_4}{\partial v_{2}}$ & $= \bar{v}_{4}\times 1$ &  $=1$  \\
    &$\bar{v}_{1}$ &$=\bar{v}_{4}\frac{\partial v_4}{\partial v_{1}}$ & $=\bar{v}_{4}\times 1$ &  $=1$  \\
    &$\bar{v}_{3}$ &$=\bar{v}_{5}\frac{\partial v_5}{\partial v_{3}}$ & $=\bar{v}_{5}\times (-1)$ &  $=-1$  \\
    &$\bar{v}_{4}$ &$=\bar{v}_{5}\frac{\partial v_5}{\partial v_{4}}$ & $= \bar{v}_{5}\times 1$ &  $=1$\\
    \cmidrule{2-5}
    \tikzmark{f}& $\bar{v}_5$ & $=\bar{y}$ & $=1$ &\\ %\bottomrule
  \end{tabular}
  \begin{tikzpicture}[overlay,remember picture, shorten >=-3pt]
\draw[->,>=triangle 60,thick] ([shift={(0ex,0ex)}]pic cs:f) -- ([shift={(0ex,0.5ex)}]pic cs:e) ;
\end{tikzpicture}
\end{center}
\end{small}
To get the desired $\bar{x}_{1}$ and $\bar{x}_2$ (i.e., $\partial y/\partial x_1$ and $\partial y/\partial x_2$), we begin with
\begin{equation*}
\bar{v}_5 =
\frac{\partial y}{\partial v_5}
=
\frac{\partial y}{\partial y} =1.
\end{equation*}
From the computational graph, we then get $\bar{v}_4$ and $\bar{v}_3$.
Because $v_4$ affects $y$ only through $v_5$, we have
\begin{equation*}
    \bar{v}_4=\frac{\partial y}{\partial v_4}
    =     \frac{\partial y}{\partial v_5}
    \frac{\partial v_5}{\partial v_4}\\
    =
    \bar{v}_5
    \frac{\partial v_5}{\partial v_4}
    = \bar{v}_5 \times 1
\end{equation*}
The above equation is based on that we already know $\partial y/\partial v_5 = \bar{v}_5$.
Also, the operation from $v_4$ to $v_5$ is an addition, so ${\partial v_5}/{\partial v_4}$ is a constant $1$.
By such a sequence, in the end, we obtain
\begin{equation*}
    \frac{\partial y}{\partial x_1} = \bar{x}_1 \text{ and }
    \frac{\partial y}{\partial x_2} = \bar{x}_2
\end{equation*}
at the same time.

\section{Implementation of Function Evaluation and the Computational Graph}
With the basic concepts ready in Section 2, we move to the implementation of the automatic differentiation.
For simplicity, we consider the forward mode. 
The reverse mode can be designed in a similar way.
Consider a function $f: R^n\to R$ with
\[
  y = f(\bx) = f(x_1, x_2, \dots, x_n).
\] 
For any given $\bx$, we show the computation of 
\[
  \dfrac{\partial y}{\partial x_1}
\]
as an example.

\subsection{The Need to Calculate Function Values}
We are calculating the derivative, so at the first glance, function values are not needed.
However, we show that function evaluation is necessary due to the function structure and the use of the chain rule.
To explain this, we begin with knowing that the function of a neural network is usually a nested composite function
\[
  f(\bx) = h_k(h_{k-1}(\dots h_1(\bx)))
\]
due to the layered structure.
For easy discussion, let us assume that $f(\bx)$ is the following general composite function
\[
  f(\bx) = g(h_1(\bx), h_2(\bx), \dots, h_k(\bx)).
\]
We can see that the example function considered earlier
\begin{equation}\label{eq:example-function}
  f(x_1, x_2) = \log x_1 + x_1x_2 - \sin x_2
\end{equation}
can be written in the following composite function
\begin{align*}
  g(h_1(x_1, x_2), h_2(x_1, x_2))
\end{align*}
with
\begin{align*}
  g(h_1, h_2) &= h_1 - h_2,\\
  h_1(x_1, x_2) &= \log x_1 + x_1x_2,\\
  h_2(x_1, x_2) &= \sin(x_2).
\end{align*}
To calculate the derivative at $\bx = \bx_0$ using the chain rule, 
we have
\[
  \dfrac{\partial f}{\partial x_1} \Big\vert_{\bx = \bx_0}
  = \sum_{i=1}^k \left(\dfrac{\partial g}{\partial h_i}
  \Big\vert_{\bh = \bh(\bx_0)}
  \times \dfrac{\partial h_i}{\partial x_1}\Big\vert_{\bx = \bx_0}\right),
\]
where the notation
\[
  \dfrac{\partial g}{\partial h_i} \Big\vert_{\bh = \bh(\bx_0)}
\]
means the derivative of $g$ with respect to $h_i$ evaluated at 
$\bh(\bx_0) 
= {\begin{bmatrix} h_1(\bx_0) & \cdots & h_k(\bx_0) \end{bmatrix}}^T$.
Clearly, we must calculate the inner function values $h_1(\bx_0), \dots, h_k(\bx_0)$ first.
The process of computing all $h_i(\bx_0)$ is part of (or almost the same as) the process of computing $f(\bx_0)$.
This explanation tells why for calculating the partial derivatives, we need the function value first.

Next, we discuss the implementation of getting the function value.
For the function~\eqref{eq:example-function}, recall we have a table recording the order to get $f(x_1, x_2)$:
\begin{center}
  \begin{tabular}{l l l l}
    \tikzmark{g}& $x_1$ &  & $=2$ \\
    & $x_2$ &  &  $=5$  \\
    \cmidrule{2-4}
    & $v_{1}$ & $= \log{x_1}$ &  $=\log{2}$  \\
    & $v_{2}$ & $= x_1\times x_2$ &  $=2\times5$  \\
    & $v_{3}$ & $= \sin{x_2}$ &  $=\sin{5}$  \\
    & $v_{4}$ & $= v_{1}+v_{2}$ &  $=0.693 + 10$  \\
    & $v_{5}$ & $= v_{4}-v_{3}$ &  $=10.693 + 0.959$  \\
    \cmidrule{2-4}
    \tikzmark{h}& $y$ & $=v_{5}$ &  $=11.652$ 
  \end{tabular}
  \begin{tikzpicture}[overlay,remember picture, shorten >=-3pt]
    \draw[->,>=triangle 60,thick] ([shift={(0ex,1.5ex)}]pic cs:g) -- (pic cs:h) ;
  \end{tikzpicture}
\end{center}
Also, we have a computational graph to generate the computing order
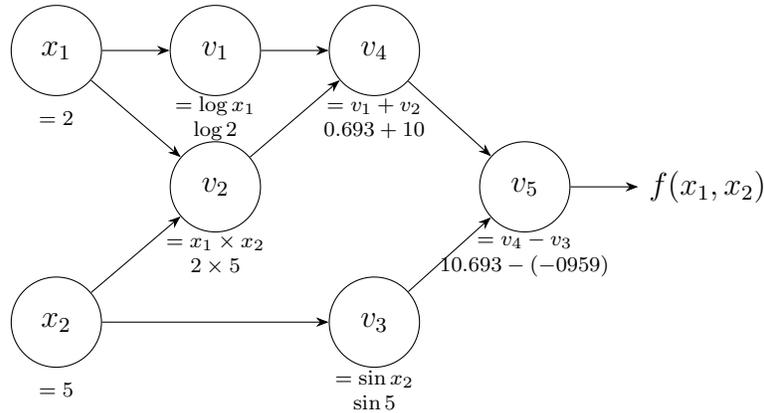
\begin{center}
  \begin{tikzpicture}[
    auto,vertex/.style={draw,circle},minimum size=1.2cm,
    every label/.append style={font=\scriptsize,below,yshift=-0.9cm}
    ]
    \node[vertex,label={$=2$}] (b) at (0,0) {$x_{1}$};
    \node[vertex,right=0.9cm of b,
    label={[align=center]$= \log x_{1}$\\$\log 2$}] (c) {$v_{1}$};
    \node[vertex,right=0.9cm of c,
    label={[align=center]$= v_1 + v_2$\\ $0.693+10$}] (d) {$v_{4}$};
    \node[vertex,below=2.4cm of b,
    label={[align=center]$= 5$}] (g) {$x_{2}$};
    \node[vertex,below=0.6cm of c,
    label={[align=center]$= x_{1}\times x_{2}$\\$2\times 5$}] (h) {$v_2$};
    \node[vertex,below=2.4cm of d,
    label={[align=center]$= \sin{x_{2}}$\\$\sin{5}$}] (i) {$v_{3}$};
    \node[vertex,right=2.9cm of h,
    label={[align=center]$= v_4 - v_3$\\$10.693-(-0959)$}] (j) {$v_5$};
    \node[right=0.9cm of j] (k) {$f(x_1, x_2)$};
    \path[-{Stealth[]}]
    (b) edge (c)
    (c) edge (d)
    (b) edge (h)
    (g) edge (h)
    (h) edge (d)
    (d) edge (j)
    (g) edge (i)
    (i) edge (j)
    (j) edge (k);
  \end{tikzpicture}
  \end{center}
Therefore, we must check how to build the graph.

\subsection{Creating the Computational Graph}
A graph consists of nodes and edges.
We must discuss what a node/edge is and how to store information.
From the graph shown above, we see that each node represents an intermediate expression:
\begin{align*}
  v_1 &= \log x_1,\\
  v_2 &= x_1 \times x_2,\\
  v_3 &= \sin x_2,\\
  v_4 &= v_{1} \times v_{2},\\
  v_5 &= v_{4} - v_{3}.
\end{align*}
The expression in each node is from an operation to expressions in other nodes.
Therefore, it is natural to construct an edge
			\[ u \to v, \]
if the expression of a node $v$ is based on the expression of another node $u$.
We say node $u$ is a parent node (of $v$) and node $v$ is a child node (of $u$).
To do the forward calculation, at node $v$ we should store $v$'s parents.
Additionally, we need to record the operator applied on the node's parents and the resulting value.
For example, the construction of the node 
			\[ v_2 = x_1 \times x_2, \]
requires to store $v_2$'s parent nodes $\{x_1,x_2\}$,
the corresponding operator ``$\times$'' and the resulting value.
Up to now, we can implement each node as a class \texttt{Node} with the following members.
\begin{center}
  \begin{tabular}{c|c|c}
    member & data type & example for \texttt{Node} $v_2$ \\
    \hline
    numerical value & \texttt{float} & $10$\\
    parent nodes & \texttt{List[Node]} & $[x_1, x_2]$\\
    child nodes & \texttt{List[Node]} & $[v_{4}]$\\
    operator & \texttt{string} & \texttt{"mul"} (for $\times$)
  \end{tabular}
\end{center}
At this moment, it is unclear why we should store child nodes in our {\tt Node} class.
Later we will explain why such information is needed.
Once the {\tt Node} class is ready, starting from initial nodes (which represent $x_i$'s), we use nested function calls to build the whole graph.
In our case, the graph for $y = f(x_1, x_2)$ can be constructed via
\begin{center}
  \begin{BVerbatim}
    y = sub(add(log(x1), mul(x1, x2)),sin(x2)).
  \end{BVerbatim}  
\end{center}
let us see this process step by step and check what each function must do.
First, our starting point is the root nodes created by the \texttt{Node} class constructor.
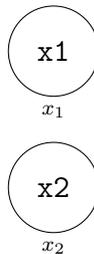
\begin{center}
  \begin{tikzpicture}[
    auto,vertex/.style={draw,circle},minimum size=1.2cm,
    every label/.append style={font=\scriptsize,below,yshift=-0.8cm}
    ]
    \node[vertex,label=\texttt{$x_1$}] (b) at (0,0) {\texttt{x1}};
    \node[vertex,below=0.6cm of b,label=\texttt{$x_2$}] (g) {\texttt{x2}};
    \path[-{Stealth[]}]
    ;
  \end{tikzpicture}
\end{center}
These root \texttt{Nodes} have empty members ``parent nodes,'' ``child nodes,'' and ``operator'' with only ``numerical value'' respectively set to $x_1$ and $x_2$.
Then, we apply our implemented \texttt{log(node)} to the node \texttt{x1}.
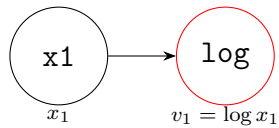
\begin{center}
  \begin{tikzpicture}[
    auto,vertex/.style={draw,circle},minimum size=1.3cm,
    every label/.append style={font=\scriptsize,below,yshift=-0.8cm}
    ]
    \node[vertex, label=\texttt{$x_{1}$}] (b) at (0,0) {\texttt{x1}};
    \node[vertex,right=0.9cm of b, draw=red,
      label=\texttt{$v_{1} = \log x_1$}] (c) {\texttt{log}};
    \path[-{Stealth[]}]
      (b) edge (c)
    ;
  \end{tikzpicture}
  \end{center}
The implementation of our \texttt{log} function should create a \texttt{Node} instance to store $\log(x_1)$.
Therefore, what we have is a wrapping function that does more than the $\log$ operation; see details in Section 3.3. The created node is the $v_1$ node in our computational graph.
Next, we discuss details of the node creation.
From the current $\log$ function and the input node $x_1$, we know contents of the following members:
\begin{itemize}
  \setlength\itemsep{-0.5em}
  \item parent nodes: $[x_1]$
  \item operator: {\tt "log"}
  \item numerical value: $\log2$
\end{itemize}
However, we have no information about children of this node.
The reason is obvious because we have not had a graph including its child nodes yet.
Instead, we leave this member ``child nodes'' empty and let child nodes to write back the information.
By this idea, our \texttt{log} function should add $v_1$ to the ``child nodes'' of $x_1$.
See more details later in Section 3.3.

We move on to apply \texttt{mul(node1, node2)} on nodes \texttt{x1} and \texttt{x2}.
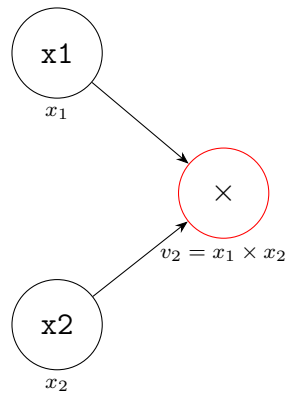
\begin{center}
  \begin{tikzpicture}[
    auto,vertex/.style={draw,circle},minimum size=1.2cm,
    every label/.append style={font=\scriptsize,below,yshift=-0.8cm}
    ]
    \node[vertex, label=$x_{1}$] (b) at (0,0) {\texttt{x1}};
    \node[vertex,below=2.4cm of b, label=$x_{2}$] (g) {\texttt{x2}};
    \node[vertex,below=0.6cm of c, draw=red, align=right,
      label=\texttt{$v_{2} = x_1 \times x_2$}] (h) {\texttt{$\times$}};
    \path[-{Stealth[]}]
    (b) edge (h)
    (g) edge (h)
    ;
  \end{tikzpicture}
  \end{center}
Similarly, the \texttt{mul} function generates a \texttt{Node} instance. 
However, different from \texttt{log(x1)}, the node created here stores two parents (instead of one). 
Then we apply the function call \texttt{add(log(x1), mul(x1, x2))}.
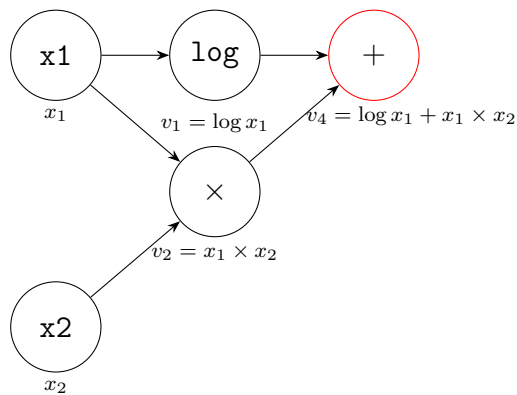
\begin{center}
  \begin{tikzpicture}[
    auto,vertex/.style={draw,circle},minimum size=1.2cm,
    every label/.append style={font=\scriptsize,below,yshift=-0.8cm}
    ]
    \node[vertex, label=$x_{1}$] (b) at (0,0) {\texttt{x1}};
    \node[vertex,below=2.4cm of b,
      label=$x_{2}$] (g) {\texttt{x2}};
    \node[vertex,right=0.9cm of b,
      label={[yshift=-0.1cm]$v_{1} = \log x_1$}] (c) {\texttt{log}};
    \node[vertex,below=0.6cm of c,
      label={$v_{2} = x_1 \times x_2$}] (h) {\texttt{$\times$}};
    \node[vertex,right=0.9cm of c, draw=red,
      label={[xshift=0.5cm]$v_{4} = \log x_1+x_1\times x_2$}] 
      (d) {\texttt{$+$}};
    \path[-{Stealth[]}]

    (c) edge (d)
    (h) edge (d)
    (b) edge (c)
    (b) edge (h)
    (g) edge (h)
    ;
  \end{tikzpicture}
  \end{center}
Next, we apply \texttt{sin(node)} to \texttt{x2}.
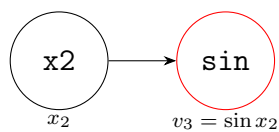
\begin{center}
  \begin{tikzpicture}[
    auto,vertex/.style={draw,circle},minimum size=1.3cm,
    every label/.append style={font=\scriptsize,below,yshift=-0.8cm}
    ]
    \node[vertex,label=$x_{2}$] (b) at (0,0) {\texttt{x2}};
    \node[vertex,right=0.9cm of b, draw=red,
      label={$v_{3} = \sin x_2$}] (c) {\texttt{sin}};
    \path[-{Stealth[]}]
      (b) edge (c)
    ;
  \end{tikzpicture}
  \end{center}
Last, applying \texttt{sub(node1,node2)} to the output nodes of \texttt{add(log(x1), mul(x1, x2))} and \texttt{sin(x1)} leads to 
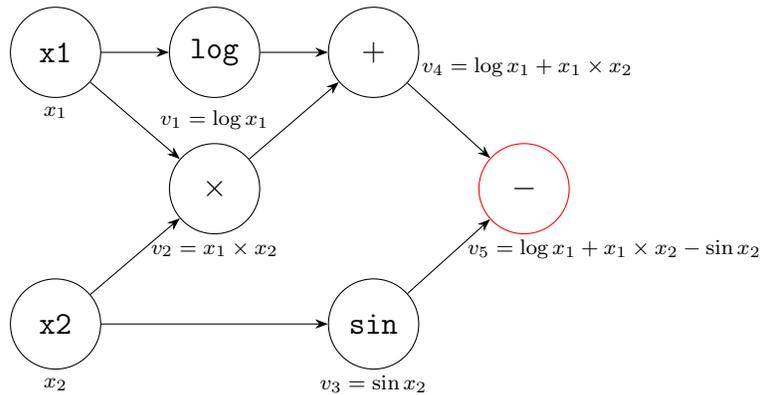
\begin{center}
  \begin{tikzpicture}[
    auto,vertex/.style={draw,circle},minimum size=1.2cm,
    every label/.append style={font=\scriptsize,below,yshift=-0.8cm}
    ]
    \node[vertex,label=$x_{1}$] (b) at (0,0) {\texttt{x1}};
    \node[vertex,below=2.4cm of b,label=$x_{2}$] (g) {\texttt{x2}};
    \node[vertex,right=0.9cm of b,
      label={[yshift=-0.1cm]$v_{1} = \log x_1$}] (c) {\texttt{log}};
    \node[vertex,below=0.6cm of c,
      label={$v_{2} = x_1 \times x_2$}] (h) {\texttt{$\times$}};
    \node[vertex,right=0.9cm of c,
      label={[right,xshift=0.5cm]$v_{4} = \log x_1+x_1\times x_2$}] (d) {\texttt{$+$}};
    \node[vertex,below=2.4cm of d,
      label={$v_{3} = \sin x_2$}] (i) {\texttt{sin}};
    \node[vertex,right=2.9cm of h,draw=red,
      label={[xshift=1.2cm]$v_{5} = \log x_1+x_1\times x_2-\sin x_2$}] (j) {$-$};
    \path[-{Stealth[]}]
    (d) edge (j)
    (i) edge (j)
    (c) edge (d)
    (h) edge (d)
    (b) edge (c)
    (b) edge (h)
    (g) edge (h)
    (g) edge (i)
    ;
  \end{tikzpicture}
\end{center}
We can conclude that each function generates exactly one \texttt{Node} instance;
however, the generated nodes differ in the operator, the number of parents, etc.

\subsection{Wrapping Functions}
We mentioned that a function like ``\texttt{mul}'' does more than calculating the product of two numbers. Here we show more details.
These customized functions ``\texttt{add}'', ``\texttt{mul}'' and ``\texttt{log}'' in the previous pages are {\it wrapping} functions, which ``wrap'' numerical operations with additional codes.
An important task is to maintain the relation between the constructed node and its parents/children.
This way, the information of graph can be preserved.

For example, we may implement the following ``\texttt{mul}'' function.
\begin{lstlisting}[caption={The wrapping function ``\texttt{mul}''},label={codeaddnode},language=Python, linerange={54-57,59-61}]
def mul(node1, node2):
    value = node1.value * node2.value
    parent_nodes = [node1, node2]
    newNode = Node(value, parent_nodes, "mul")
    node1.child_nodes.append(newNode)
    node2.child_nodes.append(newNode)
    return newNode
\end{lstlisting}
In this code, we add the created node to the ``child nodes'' lists of the two input nodes: {\tt node1} and {\tt node2}.
As we mentioned earlier, when {\tt node1} and {\tt node2} were created, their lists of child nodes were unknown and left empty.
Thus, in creating each node, we append the node to the list of its parent(s).

The output of the function should be the created node.
This setting enables the nested function call.
That is, calling 
\begin{center}
  {\tt y = sub(add(log(x1), mul(x1, x2)),sin(x2))}
\end{center}
finishes the function evaluation.
At the same time, we build the computational graph.

\section{Topological Order and Partial Derivatives}
Once the computational graph is built, we want to use the information in the graph to compute
\[
  \dfrac{\partial y}{\partial x_1}=\dfrac{\partial v_5}{\partial x_1}.
\]
\subsection{Finding the Topological Order}
\label{sec:find-topol-order}
Recall that $\partial v / \partial x_1$ is denoted by 
      $\dot{v}$.
From the chain rule,
			\begin{equation}
				\label{eq:dependency-of-v5}
				\dot{v}_5
				= \dfrac{\partial v_5}{\partial v_4}\dot{v}_4
				+ \dfrac{\partial v_5}{\partial v_3}\dot{v}_3.
			\end{equation}
We are able to calculate
\begin{equation}
  \label{eq:gradient-of-v5}
				\dfrac{\partial v_5}{\partial v_4} \text{\ \  and\ \ }
				\dfrac{\partial v_5}{\partial v_3},
	\end{equation}
because the node $v_5$ stores the needed information related to its parents $v_4$ and $v_3$.
We defer the details on calculating \eqref{eq:gradient-of-v5}, so the focus is now on calculating $\dot{v}_4$ and $\dot{v}_3$.
For $\dot{v}_4$, we further have
\begin{equation}
  \label{eq:dependency-of-v4}
  \dot{v}_4 = \dfrac{\partial v_4}{\partial v_1}\dot{v}_1 + \dfrac{\partial v_4}{\partial v_2}\dot{v}_2,
\end{equation}
which, by the same reason, indicates the need of $\dot{v}_1$ and $\dot{v}_2$.
On the other hand, we have $\dot{v}_3 = 0$ since $v_3$ (i.e., $\sin(x_2)$)
is not a function of $x_1$.
The discussion on calculating $\dot{v}_4$ and $\dot{v}_3$ leads us to find that
\begin{equation}
  \label{eq:nonreachable}
  \text{$v$ is not reachable from $x_1$ in the graph $\Rightarrow$ $\dot{v} = 0$}.
\end{equation}
We say a node $v$ is reachable from a node $u$ if there exists a path from $u$ to $v$ in the graph.
From \eqref{eq:nonreachable}, now we only care about nodes reachable from $x_1$.
Further, we must properly order nodes reachable from $x_1$ so that, for example, in \eqref{eq:dependency-of-v5}, $\dot{v}_4$ and $\dot{v}_3$ are ready before calculating $\dot{v}_5$.
Similarly, $\dot{v}_1$ and $\dot{v}_2$ should be available when calculating $\dot{v}_4$.

To consider nodes reachable from $x_1$, from the whole computational graph $G = \langle V, E\rangle$, where $V$ and $E$ are respectively sets of nodes and edges, we define
\[
  V_R = \{ v\in V \mid v \text{ is reachable from } x_1 \}
\]
and
\[
  E_R = \{ (u, v) \in E \mid u \in V_R,\ v\in V_R \}.
\]
Then,
\[
  G_R \equiv \langle V_R, E_R\rangle
\] 
is a subgraph of $G$.
For our example, $G_R$ is the following subgraph with
\[
  V_R = \{ x_1, v_1, v_2, v_4, v_5 \}
\]
and
\[
  E_R = \{(x_1, v_1), (x_2, v_2), (v_1, v_4), (v_2, v_4), (v_4, v_5)\}.
\]
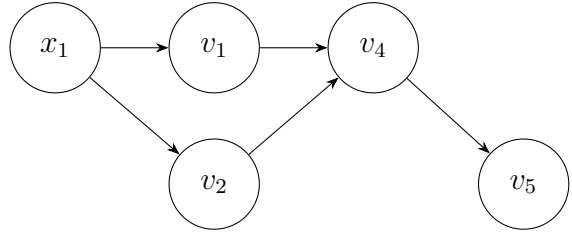
\begin{center}
  \begin{tikzpicture}[auto,vertex/.style={draw,circle},minimum size=1.2cm]
  \node[vertex] (b) at (0,0) {$x_1$};
  \node[vertex,right=0.9cm of b] (c) {$v_{1}$};
  \node[vertex,right=0.9cm of c] (d) {$v_{4}$};
  % \node[vertex,below=2.4cm of b] (g) {$x_2$};
  \node[vertex,below=0.6cm of c] (h) {$v_2$};
  % \node[vertex,below=2.4cm of d] (i) {$v_{3}$};
  \node[vertex,right=2.9cm of h] (j) {$v_5$};
  % \node[right=0.9cm of j] (k) {$y=f(x_1, x_2)$};
  \path[-{Stealth[]}]
  (b) edge (c)
  (c) edge (d)
  (b) edge (h)
  % (g) edge (h)
  (h) edge (d)
  (d) edge (j)
  % (g) edge (i)
  % (i) edge (j)
  % (j) edge (k)
  ;
  \end{tikzpicture}
\end{center}
We aim to find a ``suitable'' ordering of $V_R$ satisfying that each node $u \in V_R$ comes before all of its child nodes in the ordering.
By doing so, $\dot{u}$ can be used in the derivative calculation of its child nodes; see \eqref{eq:dependency-of-v4}.
For our example, a ``suitable'' ordering can be
\[
  x_1,v_{1},v_{2},v_{4},v_{5}.
\]
In graph theory, such an ordering is called a {\it topological ordering} of $G_R$.
Since $G_R$ is a directed acyclic graph (DAG), a topological ordering must exist.\footnote{We do not get into details, but a proof can be found in \cite{JK05a}.}
We may use depth first search (DFS) to traverse $G_R$ to find the topological ordering.
For the implementation, earlier we included a member ``child nodes'' in the {\tt Node} class, but did not explain why. The reason is that to traverse $G_R$ from $x_1$, we must access children of each node.

Based on the above idea, we can have the following code to find a topological ordering.
\begin{lstlisting}[caption={Using depth first search to find a topological ordering},label={codetporder},language=python,linerange={82-91}]
def topological_order(rootNode):
    def add_children(node):
        if node not in visited:
            visited.add(node)
            for child in node.child_nodes:
                add_children(child)
            ordering.append(node)
    ordering, visited = [], set()
    add_children(rootNode)
    return list(reversed(ordering))
\end{lstlisting}
The function \texttt{add\_children} implements the depth-first-search of a DAG.
From the input node, it sequentially calls itself by using each child as the input.
This way explores all nodes reachable from the input node.
After that, we append the input node to the end of the output list.
Also, we must maintain a set of visited nodes to ensure that each node is included in the ordering exactly once.
For our example, the input in calling the above function is $x_1$, which is also the root node of $G_R$.
The left-most path of the depth-first search is
\begin{equation}
  \label{eq:dfs-path}
  x_1\rightarrow v_1 \rightarrow v_4 \rightarrow v_5,
\end{equation}
so $v_5$ is added first. In the end, we get the following list
\[ [v_5, v_4, v_1, v_2, x_1]. \]
  % \item A node is added to the end of the \texttt{ordering}
  % list once its children were all added into the list
Then, by reversing the list, a node always comes before its children.
One may wonder whether we can add a node to the list before adding its child nodes.
This way, we have a simpler implementation without needing to reverse the list in the end.
However, this setting may fail to generate a topological ordering.
We obtain the following list for our example:
\[ [x_1, v_1, v_4, v_5, v_2].\]
A violation occurs because $v_2$ does not appear before its child $v_4$.
The key reason is that in a DFS path, a node may point to another node that was added earlier through a different path.
Then this node becomes after one of its children.
For our example,
\[ x_1 \to v_2 \to v_4 \to v_5 \]
is a path processed after the path in \eqref{eq:dfs-path}.
Thus, $v_2$ is added after $v_4$ and a violation occurs.
Reversing the list can resolve the problem.
To this end, in the function {\tt add\_children}, we must append the input node in the end.
\par In automatic differentiation, methods based on the topological ordering are called {\it tape-based} methods.
They are used in some real-world implementations such as Tensorflow.
The ordering is regarded as a tape. 
We  read the nodes one by one from the beginning of the sequence (tape) to calculate the derivative value.

Based on the obtained ordering, subsequently let us see how to compute each $\dot{v}$.
\subsection{Computing the Partial Derivative}
Earlier, by the chain rule, we have
\begin{equation*}
	\label{eq:dependency-of-v5}
	\dot{v}_5
	= \dfrac{\partial v_5}{\partial v_4}\dot{v}_4
	+ \dfrac{\partial v_5}{\partial v_3}\dot{v}_3.
      \end{equation*}
      In Section \ref{sec:find-topol-order}, we mentioned that
      \begin{equation*}
        \dot{v}_4
        \text{ and }
\dot{v}_3
\end{equation*}
should be ready before calculating $\dot{v}_5$. For
\begin{equation*}
 \dfrac{\partial v_5}{\partial v_4} \text{ and }
 \dfrac{\partial v_5}{\partial v_3},
\end{equation*}
we are able to calculate and store them when $v_5$ is
created.
The reason is that from
\begin{equation*}
  v_5(v_4, v_3) = v_4 - v_3,
\end{equation*}
we know
\begin{equation*}
  \frac{\partial v_5}{\partial v_4} = 1
  \text{ and }
  \frac{\partial v_5}{\partial v_3} = -1.
\end{equation*}

A general form of our calculation is
  \begin{equation}
    \label{eq:chain-rule-for-dot-v}
    \dot{v} = 
    \sum_{u\in \text{$v$'s parents}} 
    \dfrac{\partial v}{\partial u}\ \dot{u}.
  \end{equation}
  The second term, $\dot{u} = \dfrac{\partial u}{\partial x_1}$, comes from past calculation
  due to the topological ordering. 
  We can calculate the first term because $u$ is one of
  $v$'s parent(s) and we know the operation at $v$.
For example, we have $v_4 = v_1 \times v_2$, so
\[ 
  \dfrac{\partial v_4}{\partial v_1} = v_2
  \text{ and } 
  \dfrac{\partial v_4}{\partial v_2} = v_1.
\]
These values can be immediately computed and stored when we construct the computational graph.
Therefore, we add one member ``gradient~w.r.t.~parents'' to our \texttt{Node} class.
In addition, we need a member ``partial derivative'' to store the accumulated sum in the calculation of \eqref{eq:chain-rule-for-dot-v}.
In the end, this member stores the $\dot{v}_i$ value.
Details of the derivative evaluation are in Listing 4.
The complete list of members of our node class is in the following table.
\begin{center}
  \begin{tabular}{c|c|c}
    member & data type & 
    \begin{tabular}{@{}c@{}}
      example for \texttt{Node} $v_2$
    \end{tabular}
      \\
    \hline
    numerical value & \texttt{float} & $10$\\
    parent nodes & \texttt{List[Node]} & $[x_1, x_2]$\\
    child nodes & \texttt{List[Node]} & $[v_{4}]$\\
    operator & \texttt{string} & \texttt{"mul"}\\
    gradient w.r.t parents &\texttt{List[float]} &  $[5, 2]$\\
    partial derivative & \texttt{float} &  $5$
  \end{tabular}
\end{center}
We update the \texttt{mul} function accordingly.
\begin{lstlisting}[caption={The wrapping function ``\texttt{mul}''. The change from Listing 1 is in red color.},language=python,linerange={54-61},linebackgroundcolor={\if \value{lstnumber}\color{backcolour}\fi,\ifnum\value{lstnumber}=5\color{red}\fi}]
def mul(node1, node2):
    value = node1.value * node2.value
    parent_nodes = [node1, node2]
    newNode = Node(value, parent_nodes, "mul")
    newNode.grad_wrt_parents = [node2.value,node1.value]
    node1.child_nodes.append(newNode)
    node2.child_nodes.append(newNode)
    return newNode
\end{lstlisting}
As shown above, we must compute 
\[
  \dfrac{\partial \texttt{ newNode}}{\partial \texttt{ parentNode}}
\]
for each parent node in constructing a new child node.
Here are some examples other than the \texttt{mul} function:
\begin{itemize}
  \item For \texttt{add(node1, node2)}, we have
  \[
    \dfrac{\partial \texttt{ newNode}}{\partial \texttt{ Node1}}
    = \dfrac{\partial \texttt{ newNode}}{\partial \texttt{ Node2}} 
    = 1,
  \]
so the red line is replaced by
  \[
    \texttt{newNode.grad\_wrt\_parents = [1., 1.]}.
  \]
  \item For \texttt{log(node)}, we have 
  \[ 
    \dfrac{\partial \texttt{ newNode}}{\partial \texttt{ Node}} 
    = \dfrac{1}{\texttt{Node.value}},
  \]
  so the red line becomes
  \[
    \texttt{newNode.grad\_wrt\_parents = [1/node.value]}.
  \]
\end{itemize}
Now, we know how to get each term in \eqref{eq:chain-rule-for-dot-v}, i.e., the chain rule for calculating $\dot{v}$.
Therefore, if we follow the topological ordering, all $\dot{v}_i$ (i.e., partial derivatives with respect to $x_1$) can be calculated.
An implementation to compute the partial derivatives is as follows.
Here we store the resulting value in the member {\tt partial\_derivative} of each node.
\begin{lstlisting}[caption={Evaluating derivatives},label={codetporder},language=python,linerange={94-103}]
def forward(rootNode):
    rootNode.partial_derivative = 1
    ordering = topological_order(rootNode)
    for node in ordering[1:]:
        partial_derivative = 0
        for i in range(len(node.parent_nodes)):
            dnode_dparent = node.grad_wrt_parents[i]
            dparent_droot = node.parent_nodes[i].partial_derivative
            partial_derivative += dnode_dparent * dparent_droot
        node.partial_derivative = partial_derivative
\end{lstlisting}

\subsection{Summary}

The procedure for forward mode includes three steps:
  \begin{enumerate}
    \item Create the computational graph
    \item Find a topological order of the graph associated with $x_1$
    \item Compute the partial derivative with respect to $x_1$ along the topological order
  \end{enumerate}
We discuss not only how to run each step but also what information we should store.
This is a minimal implementation to demonstrate the forward mode automatic differentiation.

\section*{Acknowledgements}
This work was supported by National Science and Technology
Council of Taiwan grant 110-2221-E-002-115-MY3.
%---------------------bibliography

\bibliographystyle{abbrvnat}
\bibliography{autodiff}

\end{document}